\documentclass{article}

\usepackage[eandd, final]{neurips_2026}

\usepackage[utf8]{inputenc}
\usepackage[T1]{fontenc}
\usepackage{microtype}
\usepackage{hyperref}
\usepackage{url}
\usepackage{booktabs}
\usepackage{multirow}
\usepackage{graphicx}
\usepackage{amsmath}
\usepackage{amsfonts}
\usepackage{nicefrac}
\usepackage{xcolor}
\usepackage{pifont}
\usepackage{enumitem}
\usepackage{fontawesome5}

\definecolor{darkblue}{rgb}{0, 0, 0.5}
\hypersetup{colorlinks=true, citecolor=darkblue, linkcolor=darkblue, urlcolor=darkblue}

\definecolor{markyes}{HTML}{4E9A85}
\definecolor{markpart}{HTML}{DDA15A}
\definecolor{markno}{HTML}{B9BEC4}
\providecommand{\cmark}{\textcolor{markyes}{\scalebox{0.8}{\ding{51}}}}
\providecommand{\xmark}{\textcolor{markno}{\scalebox{0.8}{\ding{55}}}}
\providecommand{\partialmark}{\textcolor{markpart}{\raisebox{0.2ex}{\scalebox{0.55}{\ding{108}}}}}

\newcommand{\hficon}{\raisebox{-0.2em}{\includegraphics[height=1.1em]{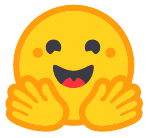}}}

\setlist[itemize]{leftmargin=1em, itemsep=-0.2em, topsep=0em}

\title{TraceML: What Auto-Research Agents Miss in Long-Horizon ML Development}

\author{%
  Jiarui Yan\thanks{Equal contribution.} \quad
  Weiwei Sun\footnotemark[1] \quad
  Sijie Li \quad
  Wenhan Li \quad
  Yiming Yang \\
  Carnegie Mellon University \\
  \texttt{\{jerryy2,yiming\}@cs.cmu.edu} \\[0.8em]
  \normalfont
  \hficon\ \href{https://huggingface.co/datasets/jerryyan/TraceML}{\textbf{Dataset}} \quad
  \hficon\ \href{https://huggingface.co/jerryyan/TraceML-Labelers}{\textbf{Models}} \quad
  \faGithub\ \href{https://github.com/JerryYan123/TraceML}{\textbf{Code}} \quad
  \faGlobe\ \href{https://jerryyan123.github.io/TraceML}{\textbf{Project page}}
}

\begin{document}

\maketitle

\begin{abstract}
Auto-research agents now run machine-learning development unattended for hours, revising data pipelines, models, and validation from their own feedback, yet on most competitions they still finish below strong human competitors. Outcome-based benchmarks record this gap but not its cause, because they grade the final submission and discard the development process behind it. We introduce TraceML, which pairs human and agent work on the same competitions under one version-level schema: 4{,}465 human Kaggle trajectories across 134 competitions, seven of which are also worked by two agent scaffolds, giving 430 paired human and 207 agent trajectories. Every code version carries its score, its timestamp, and labels for the action taken, its intent, the edit size, and the score effect. Read this way, the gap becomes concrete. Experts alternate data work, validation, model changes, and ensembling, and return to approaches they had set aside. Each agent scaffold instead collapses into a narrow loop: Codex spends its steps re-weighting ensembles and tuning submissions, MLEvolve mutates its model in place, and neither pivots at the human rate nor reopens abandoned work. A short planning prompt distilled from human practice moves the behaviors it names toward the human profile and lifts scores, but the effort profile stays agent-shaped: instruction closes only the part of the gap that reduces to instructions. We release the corpus, the schema, and the labeling models, together with an open-source toolkit that turns a run from any command-line agent into a TraceML trajectory and reads it against the human cohorts.
\end{abstract}

\vspace{0.2em}
\noindent\begin{minipage}{\textwidth}
\centering
\makeatletter\def\@captype{figure}\makeatother
\includegraphics[width=0.75\textwidth]{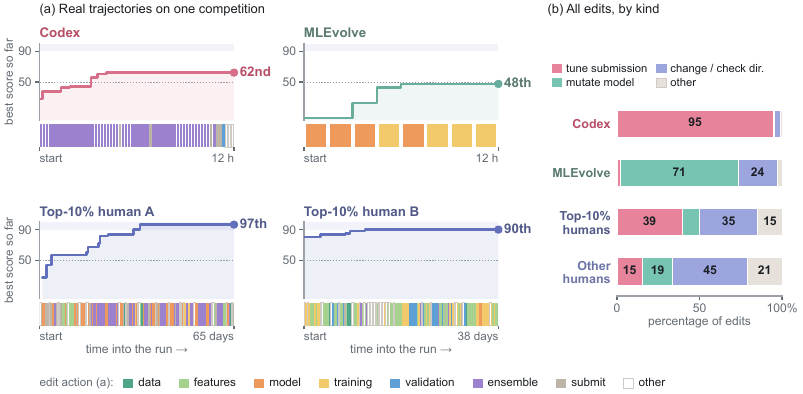}
\vspace{-0.4em}
\caption{Real trajectories on one competition. \textbf{(a)} Best score so far, as a percentile among humans, with each edit colored by its coarse action. \textbf{(b)} Share of edits by kind. Codex tunes the submission, MLEvolve mutates the model, and humans spread their edits across all kinds.}
\label{fig:teaser}
\end{minipage}
\newpage

\section{Introduction}

Large language models (LLMs) write correct code for well-specified, isolated tasks \citep{chen2021codex,lai2023ds1000}. Auto-research agents ask for more: left to run unattended, an agent must load and clean data, choose and train models, read validation signal, and revise its approach over many hours, with each decision conditioned on the last \citep{aide,Guo2024DSAgentAD,autokaggle,toledo2025aira,Liu2025MLMasterTA,li2026comind,yang2025r}. We use the term \citep{karpathy2026autoresearch,lu2024aiscientist} for any agent run this way, whether a purpose-built search scaffold or a general coding agent placed in an edit-run-evaluate loop against a grader. On Kaggle-style tasks graded through executable submissions, agents make steady progress within a run yet still finish below strong human competitors on most problems, and gain less from extra working time than humans do \citep{chan2024mlebench,wijk2024rebench}.

Outcome-based benchmarks record this gap without explaining it \citep{jing2025dsbench,huang2024mlagentbench}. They grade the final submission without seeing the sequence of edits behind it, so two runs with the same score look identical even when one experimented carefully and the other tuned blindly. Where an agent's workflow parts ways with an expert's is a question about process.

Answering it requires process-level data in a form comparable across a human and an agent working the same problem. \textbf{TraceML} represents every run as an ordered sequence of code versions, each with its leaderboard score, its timestamp, and labels for what the code contains; every transition between versions carries the action taken, its intent, the size of the edit, and its effect on the score. The same schema covers human Kaggle submissions and agent runs, so a human and an agent working the same competition can be read side by side. TraceML has 4{,}465 human trajectories from 134 Kaggle competitions; within it, a matched subset of seven competitions is worked by both humans and two auto-research agents with different search topologies, Codex and MLEvolve.

The trajectories show consistent differences (Figure~\ref{fig:teaser}). Humans alternate between exploration, diagnosis, validation, model changes, and ensembling; Codex spends most of its steps on submission-facing bookkeeping, and MLEvolve on local model and training mutations. Both scaffolds also spend more of their budget than humans do for each unit of new ground covered. A planning harness built from these observations narrows some of the behavioral differences and lifts scores on part of the competitions; the agent's effort profile stays agent-shaped.

Our contributions are: (1) \textbf{Dataset:} TraceML, a version-level trajectory dataset released alongside the extraction and labeling code, labeler checkpoints, and intervention harness. (2) \textbf{Schema:} a unified extraction and annotation schema for Kaggle notebooks, command-line interface (CLI) commits, and tree-search journals. (3) \textbf{Analysis:} process-level evidence of human-agent gaps in exploration, validation, model switching, ensembling, and repeated local optimization. (4) \textbf{Use case:} a planning harness built from these diagnostics, which narrows part of the behavioral gap and benefits the performance of the agents in some competitions. (5) \textbf{Toolkit:} an open-source command-line toolkit that extracts a run from any command-line agent, labels it with the released labeling models, and reports its development process against the human cohorts; we have applied it to five scaffolds (Appendix~\ref{app:toolkit}).

\section{Related Work}

\subsection{LLM Agents and ML-Agent Benchmarks}

LLM agents that interleave reasoning, tool use, and environment feedback \citep{yao2022react,shen2023hugginggpt} now act in web, desktop, software, and terminal environments \citep{liu2023agentbench,mialon2023gaia,Zhou2023WebArenaAR,drouin2024workarena,xie2024osworld,jimenez2024swebench,merrill2026terminalbench}, and have been applied to machine-learning development, where an agent inspects data, writes code, runs experiments, and improves on its own results \citep{aide,Guo2024DSAgentAD,autokaggle,Grosnit2024LargeLM,Hong2024DataIA,Chi2024SELATE,Trirat2024AutoMLAgentAM}. Unlike classical AutoML, which searches a fixed configuration space \citep{thornton2013auto,feurer2022auto,erickson2020autogluon}, these agents rewrite the pipeline itself, and auto-research systems carry the same loop into scientific discovery \citep{Boiko2023AutonomousCR,romera2024mathematical,lu2024aiscientist,yamada2025,tang2025airesearcher}. The two scaffolds we study differ in how they search. \textbf{Codex} is the OpenAI Codex command-line agent (\texttt{codex-cli} 0.146.0):\footnote{\url{https://github.com/openai/codex}} it runs a single edit-run-observe loop over one working directory and keeps no branch history. \textbf{MLEvolve} is an evolutionary search agent \citep{du2026mlevolve}: it grows a search tree over candidate solutions and keeps several branches alive at once. Both run on the same \texttt{gpt-5.4-mini} backend through API calls,\footnote{\url{https://developers.openai.com/api/docs/models/gpt-5.4-mini}} from the same task prompt, under the same wall-clock and GPU budget (\S\ref{sec:agent_collection}), which leaves search topology as the difference between them.

ML-agent benchmarks grade these systems on realistic tasks: MLAgentBench on bounded experimentation workflows \citep{huang2024mlagentbench}, MLE-bench on historical Kaggle competitions with held-out graders \citep{chan2024mlebench}, AIRA on the search operators and validation feedback behind MLE-bench performance \citep{toledo2025aira}, and MLGym on open-ended AI research tasks \citep{nathani2025mlgym}. Each compares an agent to other agents or to a leaderboard position, leaving no record of how a person reached the same score.

\subsection{Behavioral Diagnostics and Human-Agent Trajectories}

Many agent frameworks aim to make multi-step behavior more deliberate, among them Tree of Thoughts \citep{yao2023tree} and Reflexion \citep{shinn2023reflexion}. Evaluations of these mechanisms mostly ask whether the agent produces a valid plan in the abstract, in classical planning domains rather than ML development \citep{valmeekam2023planbench,wang2026horizon}, leaving open how such behavior plays out over hours of real ML development. Human comparison offers a reference point: RE-Bench shows that time-budgeted human baselines reveal scaling patterns that final scores hide \citep{wijk2024rebench}, and HCAST grounds autonomy evaluation in human-calibrated task attempts \citep{rein2025hcast}. Agent trajectory corpora exist for web and tool use, where they serve fine-tuning or the judging of trajectory success \citep{song2024agentbank,lu2025agentrewardbench}. None provides task-aligned human-agent trajectories for Kaggle-style ML engineering, nor a shared version-level representation of intermediate decisions. TraceML supplies both, and Table~\ref{tab:related_benchmarks} places it against related benchmarks.

\begin{table}[t]
\centering
\small
\setlength{\tabcolsep}{3pt}
\caption{Long-horizon agent benchmarks and datasets. Only TraceML keeps the code of every scored version alongside task-matched human development. Horizon: per-task agent budget (for TraceML, the span of human development). \cmark, \partialmark, \xmark: full, partial, no coverage.}
\label{tab:related_benchmarks}
\begin{tabular}{lccccc}
\toprule
Benchmark / Dataset & Score Traj & Code Traj & Human Traj & \# Envs. & Horizon \\
\midrule
MLE-bench \citep{chan2024mlebench}          & \xmark      & \xmark      & \partialmark & 75        & 24h \\
MLAgentBench \citep{huang2024mlagentbench}  & \xmark      & \partialmark & \xmark      & 13        & -- \\
AIRA \citep{toledo2025aira}                 & \partialmark & \partialmark & \xmark     & 22        & 24h \\
RE-Bench \citep{wijk2024rebench}            & \cmark      & \partialmark & \cmark      & 7         & 8h \\
HCAST \citep{rein2025hcast}                 & \xmark      & \xmark      & \cmark       & 189       & 1m--8h \\
SciCode \citep{tian2024scicode}             & \xmark      & \xmark      & \xmark       & 80 / 338  & -- \\
DiscoveryWorld \citep{jansen2024discoveryworld} & \partialmark & \xmark  & \xmark       & 120       & -- \\
HORIZON \citep{wang2026horizon}             & \partialmark & \xmark     & \xmark       & 4 domains & varies \\
TraceML (ours)                              & \cmark      & \cmark      & \cmark       & 134       & 3 weeks \\
\bottomrule
\end{tabular}
\end{table}

\section{TraceML}
\label{sec:nips_dataset}

\begin{figure}[t]
\centering
\includegraphics[width=\textwidth]{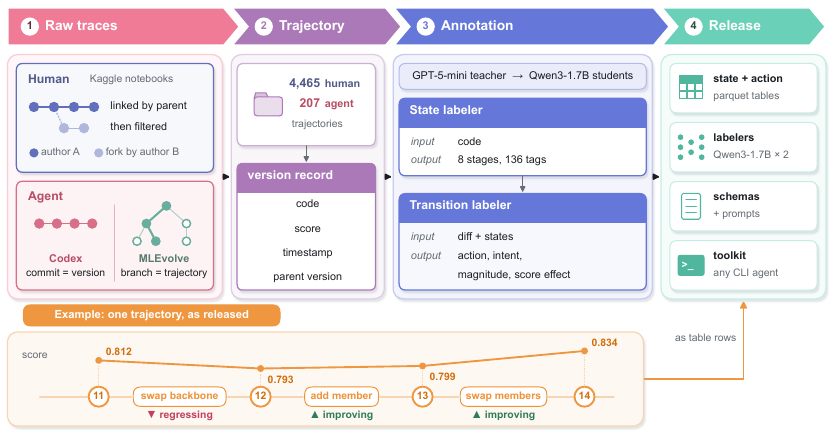}
\vspace{-0.6em}
\caption{TraceML reconstruction pipeline: notebook histories, git commits, and search branches become one version-level record. The example uses illustrative scores; Appendix~\ref{app:worked_example} follows a real trajectory.}
\label{fig:nips_substrate}
\end{figure}

Human and agent runs do not look alike at the source. A human leaves a public Kaggle notebook history built up over weeks; an agent leaves a CLI working directory or a tree-search journal produced in hours. TraceML maps both onto one representation: an ordered sequence of code versions, each with a score and a timestamp (Figure~\ref{fig:nips_substrate}). We reconstruct each side and put it on a common scoring basis (\S\ref{sec:human_collection}, \S\ref{sec:agent_collection}), annotate what each version contains and what each edit does (\S\ref{sec:dataset_schema}), and verify that those labels are reliable (\S\ref{sec:dataset_quality}).

\subsection{Human Trajectory Collection}
\label{sec:human_collection}

We reconstruct the human corpus from the Meta Kaggle database and its Code mirror\footnote{\url{https://www.kaggle.com/datasets/kaggle/meta-kaggle}, \url{https://www.kaggle.com/datasets/kaggle/meta-kaggle-code}} with a four-stage pipeline that turns raw save activity into ordered trajectories. \textbf{(1) Ingestion and alignment:} we extract public saved versions for the 134 in-scope competitions, hash each for deduplication, and join every version to its author tier and its public leaderboard score, so each trajectory carries a score at every step. \textbf{(2) Lineage reconstruction:} we recover development order as a directed acyclic graph over within-notebook histories, Kaggle fork relationships, and code-similarity links, keeping one canonical parent per version and alternate links as metadata. \textbf{(3) Pruning:} three filters drop post-deadline edits, lineages too shallow or unscored to show iteration, and score-fishing resubmissions, whose score moves without any change to the code. \textbf{(4) Normalization:} we write the surviving trajectories into the version-level format that \S\ref{sec:dataset_schema} annotates, so the schema describes each human version and transition in the same terms as its agent counterpart.

\begin{table}[t!]
\centering
\small
\setlength{\tabcolsep}{4.2pt}
\caption{TraceML corpus statistics by source. Human trajectories are broken down by Kaggle author tier. Per-version code lines is total code lines divided by snapshots within the subset.}
\label{tab:traceml_subset_summary}
\begin{tabular}{lrrrrr}
\toprule
Subset                  & Comps & Trajectories & Snapshots & Avg snap/traj & Avg lines/snap \\
\midrule
\multicolumn{6}{l}{\textit{Human (public Kaggle notebook histories)}} \\
\quad Grandmaster       & 114 &   423 & 16{,}973 & 40.1 &   555 \\
\quad Master            & 121 &   649 & 24{,}138 & 37.2 &   601 \\
\quad Expert            & 130 & 1{,}386 & 50{,}663 & 36.6 &   569 \\
\quad Contributor       & 133 & 1{,}932 & 54{,}619 & 28.3 &   494 \\
\quad Other / Unknown   &  40 &    75 &  3{,}090 & 41.2 &   295 \\
\quad \textbf{Human (all)} & \textbf{134} & \textbf{4{,}465} & \textbf{149{,}483} & \textbf{33.5} & \textbf{545} \\
\midrule
\multicolumn{6}{l}{\textit{LLM agent}} \\
\quad Codex (prior + skill)   & 7   &    18 &      579 & 32.2 & 1{,}399 \\
\quad MLEvolve         & 7   &   189 &  1{,}026 &  5.4 &   768 \\
\quad \textbf{Agent (all)} & \textbf{7} & \textbf{207} & \textbf{1{,}605} & \textbf{7.8} & \textbf{822} \\
\midrule
\textbf{Total (human $+$ agent)} & 134 & 4{,}672 & 151{,}088 & 32.3 & --- \\
\bottomrule
\end{tabular}
\end{table}

\subsection{Agent Trajectory Collection}
\label{sec:agent_collection}

TraceML pairs the human corpus with agent trajectories on seven of the competitions: 11 baseline Codex runs, 7 Codex runs carrying the planning prompt of \S\ref{sec:nips_harness}, and 13 MLEvolve searches, linearized into root-to-leaf branches (Table~\ref{tab:traceml_subset_summary}). These seven competitions define the \emph{paired subset}, where 430 human trajectories spanning all author tiers meet the agent runs under a twelve-hour agent budget; every human-agent comparison in \S\ref{sec:nips_empirical} and \S\ref{sec:nips_harness_experiment} is computed on it.

The two scaffolds leave different traces. We track Codex through sidecar Git commits on its single working directory. MLEvolve writes a search journal that gives each version one parent and keeps cross-branch reuse as separate reference edges, so we read each root-to-leaf path as a trajectory and carry a node's score to every branch through it. We re-grade every agent version with the held-out MLE-bench evaluator, not only the final submission. We then drop runs with leaky features, post-deadline data, or pretrained artifacts newer than the competition. These stages run as one command-line tool, applied so far to five scaffolds (Appendix~\ref{app:toolkit}).

\subsection{Aligning Human and Agent Trajectories}
\label{sec:alignment}
Comparing the two sources requires that a human version and an agent version denote the same kind of event. \textbf{Unit:} a human version is a Kaggle save-version, a deliberate save and not an autosave; its agent analogue is a submission-producing commit (Codex) or a search node (MLEvolve), with adjacent identical states collapsed. \textbf{Measurement:} both sides are scored by the competition's own metric. Because that metric differs across competitions, analyses relating behavior to outcome use within-competition percentile. \textbf{Observability:} public histories record saved versions, not all work, so the two sides do not reach us filtered alike. Two checks bound the effect: restricting humans to scored, submitted versions (the same event type as an agent version), and applying the human retention filters to the agent runs. Every headline gap survives both (Appendix~\ref{app:alignment_stats}). \textbf{Off-platform work} escapes either check, so we read the human corpus as a reference distribution of public practice.

\subsection{Process-Level Annotation Schema}
\label{sec:dataset_schema}
Ordering alone does not make code analyzable. The schema supplies that layer along two axes: what a version contains and what an edit does. Both vocabularies are competition-agnostic, so the schema applies to any trajectory in the same format, including scaffolds not studied here.

\textbf{Version state} assigns each version to one or more of 8 coarse ML-pipeline stages, such as feature engineering and ensembling, with a 136-tag fine vocabulary beneath them. Appendix~\ref{app:additional_empirical} shows the full schema output on a real transition (Figure~\ref{fig:nips_schema}).

\textbf{Transitions} carry most of the behavioral signal. Each is the change from one version to the next. All four labels are assigned from the diff, its surrounding code, and the grader:

\begin{itemize}[leftmargin=*, parsep=0pt, itemsep=2pt]
    \item \emph{Action:} which operations the edit performs, as a multi-label
    set over a fine vocabulary.
    \item \emph{Intent:} the purpose it serves, read from the action and the
    surrounding code.
    \item \emph{Magnitude:} how much of the working code it rewrites.
    \item \emph{Score-Effect:} whether the linked metric improves, plateaus, or
    regresses.
\end{itemize}

\subsection{Annotation Reliability and Dataset Release}
\label{sec:dataset_quality}

Hand-labeling 151{,}088 versions is not feasible, so we distill the schema into open-weight labelers: a teacher (\texttt{gpt-5.4-mini}) emits schema-constrained labels on curated traces, and those labels train two Qwen3-1.7B \citep{qwen3report} students, one for version state and one for transition action. We check the students three ways: schema compliance, agreement with an independent annotator, and a behavioral check that they reproduce expected structure (Table~\ref{tab:nips_reliability}). Agreement is lowest on intent, holds under a coarser three-class collapse that leaves every finding unchanged, and is no lower on agent transitions than on human ones. We also audit a sample stratified over human, Codex, and MLEvolve steps by hand, with the same result (Appendix~\ref{app:intent_audit}).

\textbf{Licensing and privacy.} We redistribute human notebook source only for kernels under a permissive license, verified per kernel against the Meta Kaggle Code mirror and recorded per kernel in the release. Annotations, schemas, and the pipeline code in the dataset are released under CC BY 4.0 and the toolkit under Apache 2.0; the labeler weights inherit Apache 2.0 from their Qwen3 base; Kaggle competition data itself is not redistributed. We retain author usernames and tiers because they are public Meta Kaggle metadata and the tier is the cohort variable of Table~\ref{tab:traceml_subset_summary}. Notebook execution outputs are stripped at extraction, so incidentally captured personal data does not enter the corpus. The release is documented with a datasheet \citep{gebru2018datasheets} and Croissant metadata.

\begin{figure}[t]
\centering
\includegraphics[width=\textwidth]{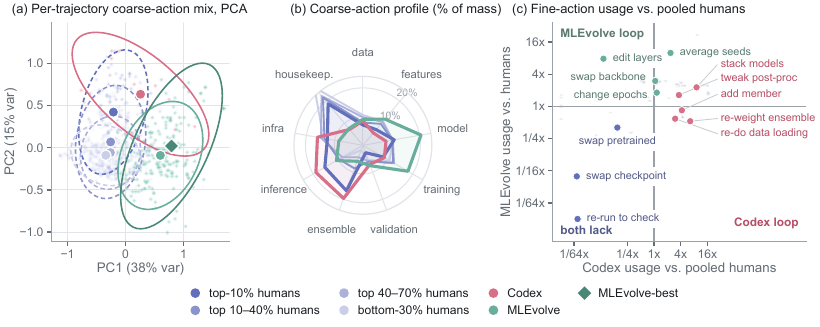}
\vspace{-1.4em}
\caption{Action profiles at two resolutions. \textbf{(a)} PCA of per-trajectory coarse-action mixes; \textbf{(b)} coarse-action profile per cohort; \textbf{(c)} fine-action usage relative to pooled humans ($\log_2$ axes). Cohorts overlap at coarse grain and separate at fine grain.}
\label{fig:overview}
\end{figure}

\section{Empirical Findings}
\label{sec:nips_empirical}

We compare Codex and MLEvolve against human leaderboard-percentile cohorts under the representation of \S\ref{sec:nips_dataset}, on the paired subset of \S\ref{sec:agent_collection} and under the same twelve-hour agent budget, with human trajectories split into a top cohort and the rest by leaderboard rank. Humans are not budget-matched and cannot be, so we read them as a reference distribution and not as a control. All intervals are clustered by run, and Appendix~\ref{app:scope} gives the per-cohort counts.

\subsection{Action Profiles: Faint at Coarse Grain, Sharp at Fine Grain}
\label{sec:nips_emp_pca}

We first ask whether human and agent trajectories occupy distinct regions of coarse-action space. Each trajectory becomes one point, the distribution of its transitions over the coarse action categories. We project those points with PCA, and we measure the distance between two cohorts as the Jensen--Shannon divergence (JSD) between their pooled action distributions (Figure~\ref{fig:overview}). The separation is partial. The four human cohorts overlap in one region while MLEvolve-best sits clearly apart, $0.09$--$0.12$ bits from every human cohort, and Codex sits about as close to top humans as the human cohorts sit to each other. Coarse action mix separates one scaffold cleanly and leaves the other indistinguishable from strong humans. The human block is also not a point, so any gap must be read against wide human variation.

Fine-grained action usage separates what the coarse mix does not (Figure~\ref{fig:overview}; tags in Appendix~\ref{app:features}). Humans spread their edits across the vocabulary, alternating data and feature work, validation, model and checkpoint changes, and ensembling, with no single tag carrying a run. Each scaffold instead settles into one band. Codex works around the submission, re-weighting ensembles, stacking models, adding members, and tweaking post-processing at several times the human rate, all of them local edits that refine a solution already in hand. MLEvolve mutates the model in place, averaging seeds, editing layers, and changing epoch counts to expand nearby variants of what it already has. What neither scaffold does is change or check direction: swapping a checkpoint, swapping a pretrained source, and re-running unchanged code to verify a result all stay an order of magnitude below the human rate.

\subsection{Agents Pivot Too Little or Too Much}
\label{sec:nips_emp_cost}

The action mix says what each cohort does; the sharper question is when a run changes direction. We define a \emph{pivot} at the action level: an edit that changes the backbone, the representation, the objective, or the validation scheme (Appendix~\ref{app:features}).

\textbf{Codex and MLEvolve miss the human rate from opposite sides.}\label{sec:nips_emp_matched} Humans pivot on $25\%$ of transitions, Codex on $9\%$, MLEvolve on $58\%$, and the contrast survives holding the code state fixed: matched to human versions in the same state, Codex is still out-pivoted three to one. \textbf{Frequent pivots do not mean good ones.} Coding each of the three steps after a pivot as improving ($+1$) or regressing ($-1$), matched humans average $+0.089$ and MLEvolve $-0.008$: its gains and losses cancel. Codex rarely turns; MLEvolve turns without gain.

\subsection{Agents Recover Scores but Not Abandoned Approaches}
\label{sec:nips_emp_memory}

\textbf{Humans return to earlier work; agents effectively never do.} A version \emph{returns} when it resembles an earlier, non-adjacent version of its own trajectory more than it resembles its predecessor, with something dissimilar reached in between, so a plateau does not qualify. Top humans return on $9\%$ of eligible versions, and $78\%$ of those returns end above the version they went back to. Across all runs Codex returns once and MLEvolve never, against the dozens the human rate predicts; a single return at that scale is chance resemblance, not a practice.

\textbf{What the agents lack is memory, not recovery.} Codex climbs back from setbacks at a rate above the top human cohort, so a run that loses ground does regain it; what it never does is reopen a line of work it had abandoned. Recovering a \emph{score} by tuning forward and returning to an earlier \emph{approach} are different capabilities, and the agents have the first without the second (Appendix~\ref{app:memory}). Together with the pivot result, this describes a search without memory: from a given state the agent does not turn, and it does not go back.

\begin{figure}[t]
\centering
\includegraphics[width=\textwidth]{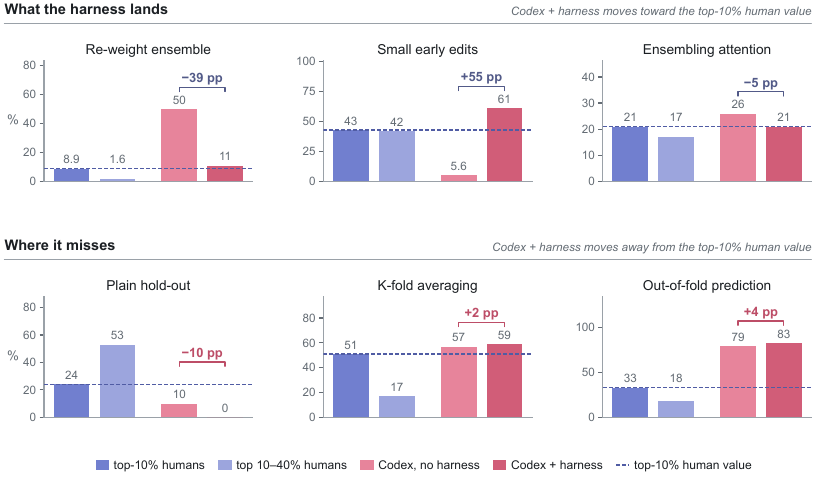}
\caption{Harness effect on six practices. Dashed line: top-10\% human value; brackets: change in Codex, in percentage points. The top three move toward the human value, the bottom three away from it.}
\label{fig:scorecard}
\end{figure}

\subsection{Agents Ensemble in Name Only}
\label{sec:nips_emp_index}

The preceding subsections describe mechanisms; this one asks which behaviors move together with final rank. Correlating all $19$ trajectory features with final leaderboard standing, on humans alone, what stands out is how a run ensembles and how large its edits are (Appendix~\ref{app:index}; associations, not causes).

\textbf{Ensembling separates work that shares a name.} All three cohorts ensemble, but $78\%$ of Codex's ensemble edits re-weight a member set it never grows, MLEvolve mostly averages seeds, and top humans put the largest share into adding a new member. Within a run, an ensemble step that adds or changes a member raises the chance the next human version improves by $6.4$ points, one that only re-weights \emph{lowers} it by $5.8$, and for Codex neither kind moves it: a checklist asking only whether the agent ensembles would rank Codex above the top human cohort while its ensemble work does nothing. \textbf{Edit size tells the same story from the other side.} Humans span the magnitude range while each scaffold works in one band: Codex edits small and pays in steps, MLEvolve edits large and pays in waste (\S\ref{sec:nips_emp_cost}). Both practices can be asked for by name, which is what makes the intervention of \S\ref{sec:nips_harness} a test rather than a guess.

\section{From Human-Agent Gaps to a Planning Harness}
\label{sec:nips_harness}

\S\ref{sec:nips_empirical} localized the human-agent gap into named behaviors. We now ask
whether naming them in a prompt changes them. The intervention is a probe: what
a prompt shifts is the part of the gap that reduces to instructions, and what
resists marks the part it does not reach.

\subsection{Harness Experiment}
\label{sec:nips_harness_experiment}

\textbf{Prompt design.}
The skill is a compact prompt block of roughly one thousand tokens that combines four mechanisms. \emph{Anti-loop constraints} prohibit the moves that sustain the mono-loop: single-holdout validation, repeated hyperparameter or post-processing tweaks, and large first-version rewrites before a working baseline exists. \emph{Human-prior practices} ask for reusable structure early: $K$-fold from the first version, an early ensemble, cached out-of-fold predictions, and multi-seed or multi-model blending. \emph{Periodic self-checks} verify every 30 to 60 minutes that validation still tracks the leaderboard and that the run has not settled into one task category. \emph{Task-specific priors} adapt the template to modality. The full prompt is in Appendix~\ref{app:prompts}.

\textbf{Experiment setup.}
We ask two questions: does the skill move behavior toward the human profile, and does the movement reach the score. Behavior is read on the trajectory features of \S\ref{sec:nips_empirical}; performance is the best valid held-out score, placed against human percentile bands so one number is comparable across scoring rules. The study covers all 7 paired competitions at the same 12-hour budget. We fix the backend (Codex CLI), the tools, the extraction pipeline, and the grader, and vary only the prompt: the \emph{baseline} arm keeps the standard task prompt and is the same run \S\ref{sec:nips_empirical} pairs against human trajectories, while the \emph{harness} arm adds the planning skill at run start and re-injects it every 30 minutes. Two further arms isolate what the skill contributes: one delivers a single content block instead of the full skill, and one keeps the re-injection cadence with the planning content removed. Repeated same-condition runs bound the noise at roughly $0.01$ in each competition's metric (Appendix~\ref{app:harness_scores}).

\subsection{Results}
\label{sec:nips_harness_results}

\textbf{Three behaviors move onto the human value} (Figure~\ref{fig:scorecard}). The agent stops re-weighting an ensemble where it should add a member, shifts attention toward ensembling, and starts making the small early edits it previously skipped almost entirely. The corrections are not marginal: re-weighting falls roughly fivefold, and small early edits rise from near zero to above the top human rate. What the three share is room to move.

\textbf{Where the prompt misses, it misses in two ways.} It overshoots what it forbids: the plain hold-out goes to zero, well under the quarter of states at which top humans still use one, because a ban gives a direction but not a destination. And it saturates on what it prescribes: Codex already ran $K$-fold averaging and persisted out-of-fold predictions at or above human rates before being asked, so requesting more of either moves nothing.

\textbf{Scores improve, and the content is what does it.} Five of the seven competitions improve, two are within noise, and none regress (Appendix~\ref{app:harness_scores}). Removing the planning content while keeping the injection schedule lands at or below the baseline everywhere, so the gain comes from what the skill says rather than how often it is repeated.

A practice transfers when the instruction names a level the agent has not already passed. It does not transfer when the instruction is a direction with no destination, which is what a prohibition is, nor when the agent already stands beyond the human value. The prompt is therefore useful less as a fix than as a probe: it marks the boundary between what instruction can reach and what it cannot, and what lies past it is changing the agent itself.

\begin{figure}[t]
\centering
\includegraphics[width=\textwidth]{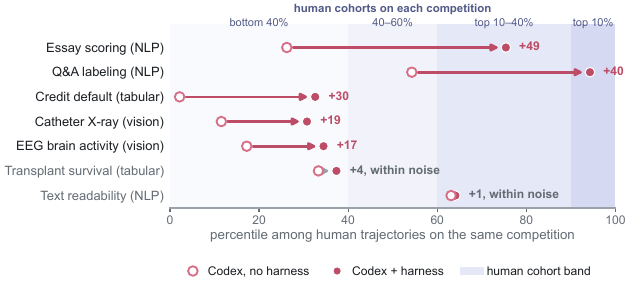}
\vspace{-1em}
\caption{Codex leaderboard percentile without (hollow) and with (filled) the harness; bands are human cohorts. Five competitions improve, two stay within noise.}
\label{fig:progress}
\end{figure}

\section{Discussion and Future Work}
\label{sec:outlook}

Reading development as a process turns ``the agent scores lower'' into specific
things the agent does not do, and two of them look like design problems rather
than model problems. The first is memory. Agents never go back to work they set
aside (\S\ref{sec:nips_emp_memory}), and the decision to go back is not what
they are missing: Codex recovers from setbacks as well as top humans, and gains
what a human gains on the rare occasions it does change direction. What it lacks
is its own history in a form it can search, which makes retrieval over a run's
earlier states a concrete thing to build. The second is control. Both scaffolds
miss the human pivot rate from opposite sides and only the human pivots pay
(\S\ref{sec:nips_emp_cost}), yet each follows one policy throughout, so what is
wanted is a controller that reads where the run stands rather than a stronger
base model. Either idea is now
cheap to test. Final scores cannot separate these cases, and neither can a
checklist of whether a practice occurred, since all three cohorts ensemble and
only the human ensembling changes anything
(\S\ref{sec:nips_emp_index}). The released pipeline turns a run from any
command-line agent into a TraceML trajectory and a report against the human
cohorts in minutes, and we have applied it to five scaffolds (Appendix~\ref{app:toolkit}). The human corpus is fixed
while agents keep changing, so it can serve as a reference that new systems are
read against as they appear rather than a benchmark that ages with them.

\section{Conclusion}

TraceML pairs human and agent ML development on the same competitions under one
version-level schema, so two runs can be compared through the work behind a
submission rather than the submission alone. The paired data shows agents and
experts developing differently, and not as one clean gap. The two scaffolds miss
the human profile from opposite sides, one tuning without changing
direction and the other changing direction without consolidating. A planning prompt built from these
diagnostics moves the behaviors that reduce to a yes or no check and leaves the
rest, which is where instruction stops and agent design begins. We release the
corpus, the schema, the labelers and the extraction pipeline, so that new
auto-research agents can be read against human practice as they appear.

\bibliographystyle{plainnat}
\bibliography{ref}

\appendix
\section{Use of LLMs}

We used large language models only to help draft and edit the text of this paper.

\section{Dataset Construction and Alignment}
\label{app:construction}

This appendix supports \S\ref{sec:human_collection}--\S\ref{sec:alignment}: what
the human retention filter removes (\ref{app:filter_audit}), how much visible
reuse the corpus contains (\ref{app:reuse}), how the human and agent units line
up (\ref{app:alignment_stats}), which runs enter which analysis
(\ref{app:scope}), the extraction toolkit (\ref{app:toolkit}), and a worked
example of the pipeline (\ref{app:worked_example}).

\subsection{Human Retention Filter: Audit of Removed Kernels}
\label{app:filter_audit}

The filter of \S\ref{sec:human_collection} verifies only that a kernel is an in-window development trajectory. Its conditions are: every version inside $[\text{launch}, \text{deadline}]$; a chain of at least 5 versions spanning at least 3 days carrying at least 1 score; and no near-static resubmission where the score changes but the code does not. No term references phase variety, action diversity, or intent. Table~\ref{tab:filter_audit} reports what it removes.

\begin{table}[h]
\centering
\small
\setlength{\tabcolsep}{6pt}
\caption{What the retention filter removes, from 5{,}048 candidate kernels. The removed set is dominated by post-deadline write-ups rather than by weak development trajectories, and its medal rate is comparable to that of the retained set, so the filter does not remove a weak tail.}
\label{tab:filter_audit}
\begin{tabular}{p{0.26\textwidth}rp{0.50\textwidth}}
\toprule
Outcome & Kernels & Character of the set \\
\midrule
Retained & 4{,}465 & 47.3\% medalled \\
Removed: no in-window\newline version & 382 & 99.2\% published entirely after the deadline, median 299 days late \\
Removed: failed content\newline conditions & 201 & Too few versions, too short a span, or no score \\
\midrule
\textbf{Removed (all)} & \textbf{583} & \textbf{43.3\% medalled} \\
\bottomrule
\end{tabular}
\end{table}

Applying the same conditions to agents (\S\ref{sec:alignment}) is possible for two of the three: the one-score condition passes for 100\% of runs in both cohorts, and the five-version condition passes for 9 of 11 Codex runs and 111 of 189 MLEvolve branches. The three-day span condition does not apply to runs measured in hours. Restricting the comparison to filter-passers leaves every gap of \S\ref{sec:nips_empirical} intact and widens Codex's coarse action-space distance from humans from $0.077$ to $0.095$ bits.

The retention rule therefore does not favor the human side.

\subsection{Fork and Reuse Rates}
\label{app:reuse}

Kaggle notebooks may share fork lineage or copy public baselines, which makes trajectory units statistically dependent and bounds how much of a ``human trajectory'' is original work. Table~\ref{tab:reuse} measures both.

\begin{table}[h]
\centering
\small
\setlength{\tabcolsep}{6pt}
\caption{Visible reuse in the retained corpus and in the paired sample used for human-agent comparison. Near-duplicate code is code similarity $\geq 0.9$ against another kernel; all observed near-duplicate links fall within the same competition, consistent with shared starter templates.}
\label{tab:reuse}
\begin{tabular}{lcc}
\toprule
Reuse measure & Full corpus (4{,}465) & Paired sample (430) \\
\midrule
Has a fork parent & 8.2\% & 8.6\% \\
Near-duplicate code in $\geq 1$ version & 23.7\% & 21.0\% \\
Near-duplicate code in a majority of versions & --- & 7.9\% \\
Near-duplicate code across the whole trajectory & --- & 3.3\% \\
\bottomrule
\end{tabular}
\end{table}

Because fork lineage and shared baselines induce correlated samples, all uncertainty estimates in \S\ref{sec:nips_empirical} cluster by competition, by human fork-lineage group, and by agent run, with MLEvolve branches that share tree nodes resampled together.

\subsection{Unit Alignment Statistics}
\label{app:alignment_stats}

Table~\ref{tab:alignment_stats} reports the alignment of the version unit described in \S\ref{sec:alignment}.

\begin{table}[h]
\centering
\small
\setlength{\tabcolsep}{6pt}
\caption{How the version unit lines up across sources. Agent versions are submission-producing commits with adjacent identical-code commits collapsed, which is the analogue of a deliberate Kaggle save-version rather than a raw log entry.}
\label{tab:alignment_stats}
\begin{tabular}{lccc}
\toprule
 & Human & Codex & MLEvolve \\
\midrule
Median versions per trajectory & 20 & 18 & 5 \\
Versions carrying a score & 45\% & $\approx$99\% & $\approx$99\% \\
Saved versions that are syntactically valid code & 97.9\% & --- & --- \\
\bottomrule
\end{tabular}
\end{table}

The collapse from raw records to versions is substantial on the agent side: in one representative Codex run, 1{,}290 graded commits reduce to 16 versions.

\subsection{Analysis Scope and Cohort Composition}
\label{app:scope}

The paper reads three nested scopes, and every count in the main text belongs to
exactly one of them. Table~\ref{tab:scope} states them. The \emph{corpus} is the
full release. The \emph{paired subset} is the 7 competitions worked by both
sides, which fixes the human reference for all agent comparisons. The
\emph{twelve-hour scope} restricts the agent side to runs sharing a budget and
covers the same 7 competitions and the same 430 human trajectories; it is where
every behavioral comparison of \S\ref{sec:nips_empirical} is computed. The
harness experiment of \S\ref{sec:nips_harness_experiment} adds the intervention
and ablation arms on the same 7.

\begin{table}[h]
\centering\small
\setlength{\tabcolsep}{5pt}
\caption{The three scopes. Human trajectories are split into a top cohort and
the rest by leaderboard rank. MLEvolve branches share tree nodes, so
the run count rather than the branch count sets the effective sample size.}
\label{tab:scope}
\begin{tabular}{lccccc}
\toprule
& & & & \multicolumn{2}{c}{MLEvolve} \\
\cmidrule(lr){5-6}
Scope & Comps & Human traj. & Codex runs & runs & branches \\
\midrule
Corpus (\S\ref{sec:human_collection})                    & 134 & 4{,}465 & ---   & ---  & --- \\
Paired subset (\S\ref{sec:agent_collection})             & 7   & 430     & 11    & 13   & 189 \\
Twelve-hour scope (\S\ref{sec:nips_empirical})           & 7   & 430     & 10    & 3    & 107 \\
Harness arms (\S\ref{sec:nips_harness_experiment})       & 7   & ---     & 30    & ---  & --- \\
\bottomrule
\end{tabular}
\end{table}

\subsubsection*{Per-run composition of the Codex cohort}

The ten twelve-hour Codex runs differ enormously in length, and the short ones
are unstable. Table~\ref{tab:codex_runs} lists every run with its marginal pivot
rate. The six runs longer than $90$ transitions all sit between $2.0\%$ and
$4.8\%$; the four shorter than $10$ transitions read $0\%$, $0\%$, $57.1\%$ and
$62.5\%$, the last two being four and five pivots respectively. Pooling
transitions across runs would let a seven-transition run speak as loudly per
observation as a thousand-transition one, so every interval in
\S\ref{sec:nips_empirical} is clustered by run, and the matched-state analysis of
\S\ref{sec:nips_emp_matched} reports a run-clustered CI for this reason.

\begin{table}[h]
\centering\small
\setlength{\tabcolsep}{6pt}
\caption{The ten twelve-hour Codex runs. ``Pivot rate'' is the marginal
(unmatched) share of transitions carrying a pivot tag. Runs are ordered by
length; the four shortest carry almost no information individually.}
\label{tab:codex_runs}
\begin{tabular}{llrr}
\toprule
Competition & Track & Transitions & Pivot rate \\
\midrule
\texttt{gquest}    & agent   & 1{,}017 & 2.1\% \\
\texttt{gquest}    & llm\_v3 &    226 & 2.2\% \\
\texttt{ranzcr}    & agent   &    166 & 4.8\% \\
\texttt{ranzcr}    & llm\_v3 &    102 & 2.0\% \\
\texttt{hms}       & agent   &     97 & 2.1\% \\
\texttt{aes2}      & llm\_v3 &     64 & 1.6\% \\
\midrule
\texttt{aes2}      & agent   &      8 & 62.5\% \\
\texttt{commonlit} & agent   &      7 & 57.1\% \\
\texttt{amex}      & agent   &      5 & 0.0\% \\
\texttt{equity}    & agent   &      3 & 0.0\% \\
\midrule
\multicolumn{2}{l}{\textbf{Pooled}} & \textbf{1{,}695} & \textbf{2.8\%} \\
\bottomrule
\end{tabular}
\end{table}

\subsection{Cross-Scaffold Extraction Toolkit}
\label{app:toolkit}

Extraction, grading, labeling, and behavior reporting are packaged as a single command that accepts a run directory from any CLI agent. We have applied it to five scaffolds: Codex CLI, MLEvolve, AIDE, Claude Code, and Gemini CLI. Each run returns a TraceML-schema trajectory and a report against the released human cohorts, in minutes on one A6000 GPU. As an illustration, a one-hour Claude Code run (\texttt{haiku-4.5} backend) on \texttt{commonlitreadabilityprize} yields 13 distinct code states and a best RMSE of $0.733$, above 52\% of the human cohort, with an exploration-heavy intent mix ($36\%$ against $8\%$ for top humans). The toolkit is what lets the human cohorts serve as a reference for scaffolds that did not exist when the corpus was built.

\subsection{Worked Example of the Pipeline}
\label{app:worked_example}

To make the schema concrete, we follow one retained trajectory end to end: a Grandmaster \texttt{commonlitreadabilityprize} kernel with 20 saved versions and therefore 19 transitions. Table~\ref{tab:worked_example} shows four consecutive transitions from its middle, where the trajectory reaches its best score.

\begin{table}[h]
\centering
\small
\setlength{\tabcolsep}{4pt}
\caption{Four consecutive transitions from one human trajectory, as represented in TraceML. Each transition carries actions, intents, a magnitude, and the score effect read from the leaderboard.}
\label{tab:worked_example}
\begin{tabular}{lp{0.27\textwidth}p{0.16\textwidth}ll}
\toprule
Transition & Actions & Intents & Magn. & Score \\
\midrule
$v_{13}\to v_{14}$ & model, training, housekeeping, infra & optimization + debugging & micro & $0.535\to0.509$ (improving) \\
$v_{14}\to v_{15}$ & data, augmentation, training, model & exploration + optimization & minor & unscored \\
$v_{15}\to v_{16}$ & data, training, validation, housekeeping & debugging + optimization & micro & $\to 0.4989$ (trajectory best) \\
$v_{16}\to v_{17}$ & training, model, infra & optimization & micro & $0.4989\to0.5062$ (regressing) \\
\bottomrule
\end{tabular}
\end{table}

Read as development decisions, the four steps are: switch to \texttt{roberta-large} and fix weight restoration before evaluation; inject target-noise sampling using the per-example standard error; fix scalar extraction of validation labels so out-of-fold evaluation is correct; and finally train longer on the base checkpoint, which regresses. The alternation of optimization with debugging, and the willingness to keep a step that did not improve the score, is the pattern \S\ref{sec:nips_empirical} finds largely absent from agent trajectories.

\section{Annotation Schema and Reliability}
\label{app:additional_empirical}

\begin{figure}[!htbp]
\centering
\includegraphics[width=\textwidth]{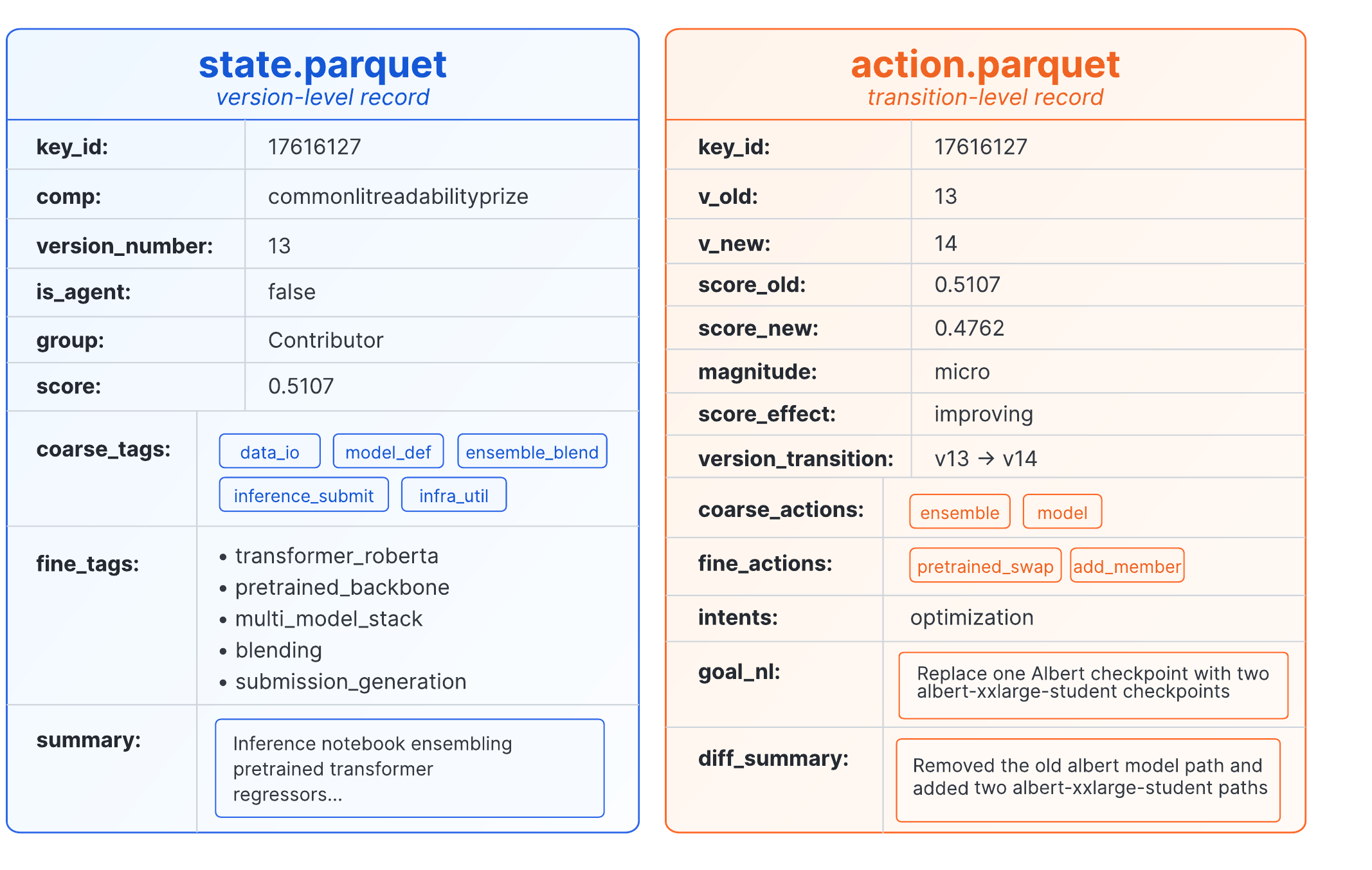}
\caption{Example schema output for one version and its transition.}
\label{fig:nips_schema}
\end{figure}

\begin{table}[h]
\centering
\small
\setlength{\tabcolsep}{6pt}
\caption{
Annotation reliability across teacher stability, cross-model agreement, and teacher--student transfer.
We report Cohen's $\kappa$ and multi-label Jaccard (J) on held-out state and action annotations; teacher--student scores compare the released Qwen3-1.7B labelers with the \texttt{gpt-5.4-mini} teacher.
}\label{tab:nips_reliability}
\begin{tabular}{lccc}
\toprule
\textbf{Annotation level (\#tags)} & \textbf{Self-consistency} & \textbf{Cross-model} & \textbf{Teacher\,$\to$\,Student} \\
\midrule
State coarse (8)       & $\kappa{=}0.872$\,/\,J{=}$0.969$ & $\kappa{=}0.801$\,/\,J{=}$0.954$ & $F_1^{\mathrm{macro}}{=}0.978$ \\
Action coarse (10)     & J{=}$0.875$                       & J{=}$0.641$                       & $F_1^{\mathrm{macro}}{=}0.733$ \\
Intent (6 classes)     & $\kappa{=}0.834$                  & $\kappa{=}0.611$                  & $\mathrm{acc}{=}0.772$         \\
Magnitude (4 levels)   & $\kappa{=}0.921$                  & $\kappa{=}0.576$                  & $\mathrm{acc}{=}0.928$         \\
\bottomrule
\end{tabular}
\end{table}

Figure~\ref{fig:nips_schema} shows the schema output on a real transition.

\subsection{Intent Label Audit by Cohort}
\label{app:intent_audit}

Table~\ref{tab:intent_audit} reports the per-cohort human audit summarized in \S\ref{sec:dataset_quality}.

Cross-model agreement between two independent LLM annotators on 499 held-out
transitions is $\kappa{=}0.611$, a conservative bound that counts unparseable
outputs as disagreements; on items where both produced a valid label,
$\kappa{=}0.724$. The 100-item human audit agrees with the released labels
$81\%$ of the time ($\kappa{=}0.68$). Collapsing the six intent classes to three
under two independent groupings changes no finding. The claim that leans
hardest on intent, that agents rarely diagnose, also holds on a label-free
proxy: explicit error-fixing edits appear in $11.1\%$ of human transitions
against $1.5\%$ of Codex's.

\begin{table}[h]
\centering
\small
\setlength{\tabcolsep}{6pt}
\caption{Human audit of released intent labels, stratified over the three cohorts. Agreement on agent transitions is not lower than on human ones, so the human-agent behavioral gaps of \S\ref{sec:nips_empirical} are not an artifact of the labels being less reliable on agent code. The $75$--$85\%$ spread lies within binomial uncertainty at these sample sizes (roughly $\pm12$ points at $n{=}40$).}
\label{tab:intent_audit}
\begin{tabular}{lccc}
\toprule
Cohort & $n$ & 6-class agreement & 3-class collapse \\
\midrule
Human    & 40  & 80.0\% & 82.5\% \\
Codex    & 40  & 85.0\% & 87.5\% \\
MLEvolve & 20  & 75.0\% & 75.0\% \\
\midrule
\textbf{Overall} & \textbf{100} & \textbf{81\%} & \textbf{83\%} ($\kappa{=}0.68$) \\
\bottomrule
\end{tabular}
\end{table}

\section{Additional Statistics for the Behavioral Findings}
\label{app:behavioral}

This appendix supports \S\ref{sec:nips_empirical}: whether the headline gaps
survive every fairness treatment (\ref{app:robustness}), the action
distributions the divergences are computed from (\ref{app:action_freq}), and the
two senses in which a run can go back to earlier work (\ref{app:memory}).

\subsection{Headline Gaps Under Every Fairness Treatment}
\label{app:robustness}

Each column perturbs one side of the comparison. \emph{Scored-only} restricts humans to scored, submitted versions, the same event type as an agent version. \emph{Agents filtered} applies the human retention conditions of \S\ref{sec:human_collection} to agents. \emph{$+$New runs} adds the matched-budget agent runs of \S\ref{sec:nips_harness_experiment}. Table~\ref{tab:robustness} reports the result. Confidence intervals come from a two-stage cluster bootstrap over competitions, then over human fork-lineage clusters and agent runs, with MLEvolve branches sharing tree nodes resampled together ($B{=}4000$).

\begin{table}[h]
\centering
\small
\setlength{\tabcolsep}{3.5pt}
\caption{Every headline gap survives every fairness treatment, and every interval excludes zero. Gaps are human minus agent, in percentage points except for the divergence rows. Note that \emph{agents filtered} \emph{widens} rather than narrows the Codex gaps.}
\label{tab:robustness}
\begin{tabular}{lccccc}
\toprule
Gap (human $-$ agent) & Orig. & Scored-only & Agents filt. & $+$New runs & 95\% cluster CI \\
\midrule
Debugging intent, vs Codex     & $+11.4$ & $+8.0$  & $+11.6$ & $+10.1$ & $[+6.0, +14.0]$ \\
Debugging intent, vs MLEvolve  & $+9.6$  & $+6.2$  & $+9.0$  & ---     & $[+7.7, +12.4]$ \\
Ensemble action mass, vs Codex & $-13.4$ & $-12.9$ & $-14.8$ & $-10.4$ & $[-19.9, -7.7]$ \\
Model$+$train.\ mass, vs MLEvolve & $-16.5$ & $-16.3$ & $-15.1$ & --- & $[-21.3, -10.9]$ \\
Action JSD (bits), vs Codex    & $0.077$ & $0.080$ & $0.095$ & $0.041$ & $[0.047, 0.145]$ \\
Action JSD (bits), vs MLEvolve & $0.054$ & $0.061$ & $0.047$ & ---     & $[0.048, 0.084]$ \\
\bottomrule
\end{tabular}
\end{table}

A transition-level generalized estimating equation (binomial, exchangeable correlation, same clusters) regressing debugging intent on cohort agrees with the bootstrap: the coefficient is $-1.82$ (SE $0.52$, $z{=}-3.5$) for Codex and $-1.89$ (SE $0.29$, $z{=}-6.6$) for MLEvolve.

\subsection{Full Coarse-Action Frequency Table}
\label{app:action_freq}

Figure~\ref{fig:overview} compares cohorts by the divergence between their
action distributions; Table~\ref{tab:action_freq} gives those distributions
themselves. Each trajectory's coarse-action histogram is normalized and then
averaged within the cohort, so every trajectory contributes equally regardless of
length. A transition may carry more than one coarse action, so a row is a share
of action mass rather than of transitions. The table is computed on the paired
subset rather than the twelve-hour scope, so it covers every human trajectory
that meets an agent.

The gaps quoted in Table~\ref{tab:robustness} read directly off this table:
ensemble mass is $6.3$ for humans against $19.7$ for Codex ($-13.4$), and
model-plus-training mass is $21.9$ against MLEvolve's $38.4$ ($-16.5$).
Table~\ref{tab:action_quintile} gives the per-quintile split for the two
headline cohorts, with each trajectory divided into five equal position
quintiles.

\begin{table}[h]
\centering\small
\setlength{\tabcolsep}{7pt}
\caption{Share of coarse-action mass per cohort, in percent, trajectory-equal.
Columns need not sum to $100$ exactly because a transition can carry several
actions. Codex concentrates in \texttt{ensemble} and \texttt{inference},
MLEvolve in \texttt{model} and \texttt{training}, and both spend roughly a third
of the human share on \texttt{housekeeping}.}
\label{tab:action_freq}
\begin{tabular}{lrrrr}
\toprule
Coarse action & Top-40\% humans & All humans & Codex & MLEvolve \\
\midrule
\texttt{data} & 8.0 & 9.1 & 8.5 & 9.0 \\
\texttt{features} & 9.1 & 10.0 & 5.6 & 11.9 \\
\texttt{augmentation} & 0.4 & 0.4 & 0.2 & 0.4 \\
\texttt{model} & 10.7 & 10.5 & 8.2 & 20.4 \\
\texttt{training} & 11.4 & 11.4 & 8.6 & 18.0 \\
\texttt{ensemble} & 9.8 & 6.3 & 19.7 & 6.0 \\
\texttt{validation} & 5.4 & 5.5 & 7.0 & 7.7 \\
\texttt{inference} & 10.3 & 9.8 & 18.9 & 10.7 \\
\texttt{infra} & 13.5 & 14.0 & 16.1 & 8.5 \\
\texttt{housekeeping} & 21.3 & 22.9 & 7.2 & 7.4 \\
\midrule
$n$ (trajectories) & 173 & 425 & 11 & 189 \\
\bottomrule
\end{tabular}
\end{table}

\begin{table}[h]
\centering\small
\setlength{\tabcolsep}{3.4pt}
\caption{Per-quintile split of the coarse-action shares of
Table~\ref{tab:action_freq}, in percent, for the top-40\% human cohort and
Codex. Q1--Q5 are within-trajectory position quintiles. Human ensembling rises
steadily over the run (7.3 to 11.3) while Codex holds its ensemble-and-inference
band from the first quintile on.}
\label{tab:action_quintile}
\begin{tabular}{llrrrrrrrrrr}
\toprule
Cohort & Quintile & \texttt{data} & \texttt{feat.} & \texttt{aug.} & \texttt{model} & \texttt{train.} & \texttt{ens.} & \texttt{valid.} & \texttt{infer.} & \texttt{infra} & \texttt{house.} \\
\midrule
\multirow{5}{*}{\shortstack[l]{Top-40\%\\humans}}
 & Q1 & 9.4 & 8.7 & 0.2 & 9.8 & 10.3 & 7.3 & 4.1 & 9.6 & 15.1 & 25.4 \\
 & Q2 & 7.8 & 8.7 & 0.6 & 10.4 & 13.5 & 9.1 & 4.9 & 9.2 & 13.5 & 22.4 \\
 & Q3 & 7.7 & 9.0 & 0.3 & 10.8 & 10.6 & 10.1 & 5.9 & 10.3 & 13.0 & 22.2 \\
 & Q4 & 6.6 & 9.1 & 0.4 & 9.4 & 11.1 & 10.5 & 5.0 & 10.8 & 12.5 & 24.5 \\
 & Q5 & 6.5 & 9.3 & 0.3 & 10.6 & 11.3 & 11.3 & 5.2 & 10.9 & 12.0 & 22.6 \\
\midrule
\multirow{5}{*}{Codex}
 & Q1 & 8.7 & 5.4 & 0.2 & 6.3 & 8.6 & 22.6 & 9.7 & 23.9 & 11.0 & 3.4 \\
 & Q2 & 5.5 & 5.8 & 0.0 & 8.7 & 11.9 & 20.0 & 8.9 & 19.4 & 14.6 & 5.2 \\
 & Q3 & 8.5 & 5.1 & 0.5 & 9.6 & 10.7 & 21.6 & 6.2 & 18.2 & 15.4 & 4.1 \\
 & Q4 & 8.2 & 6.8 & 0.0 & 9.9 & 8.1 & 20.8 & 5.9 & 20.0 & 15.0 & 5.3 \\
 & Q5 & 9.3 & 5.5 & 0.0 & 7.1 & 7.4 & 17.1 & 5.7 & 18.0 & 18.8 & 11.1 \\
\bottomrule
\end{tabular}
\end{table}

\subsection{Returning to Earlier Work}
\label{app:memory}

\S\ref{sec:nips_emp_memory} reports that agents do not go back, and
\S\ref{sec:outlook} treats that as the clearest mechanism the data names.
Table~\ref{tab:memory} gives the two measurements behind those statements,
computed by the released \texttt{solution\_revisit.py} and
\texttt{setback\_recovery.py} on the twelve-hour scope of
Appendix~\ref{app:scope}. A version is eligible once its trajectory has at least
four versions and it is not among the first two.

The two rows measure different things. A \emph{solution revisit} asks whether a
run returns to an earlier \emph{approach}: a version qualifies when its state
signature resembles an earlier non-adjacent version more than it resembles its
own predecessor, and the run reached something dissimilar in between, so a
plateau does not count. A \emph{setback recovery} asks only whether a run that
fell below its own best \emph{score} climbs back, which a purely local edit can
achieve without any return to earlier work. Agents recover scores at or above
the human rate (Codex recovers $89\%$ of its setbacks) while almost never
revisiting a solution.

\begin{table}[h]
\centering\small
\setlength{\tabcolsep}{5pt}
\caption{Two senses of going back, on the twelve-hour scope. Solution revisits
use the released state signature (Jaccard $\ge 0.6$, closer to the earlier
version than to the predecessor by $0.1$, with the run reaching something below
$0.5$ similarity in between); setback recovery uses scores only. Codex's single
revisit in $658$ eligible versions and MLEvolve's zero in $344$ are against the
$60$ and $31$ occurrences the top human rate would predict.}
\label{tab:memory}
\begin{tabular}{lrrrr}
\toprule
 & Top humans & Other humans & Codex & MLEvolve \\
\midrule
\multicolumn{5}{l}{\emph{Solution revisit (state signature)}} \\
\quad \% of eligible versions       & 9.1\% & 7.2\% & 0.2\% & 0.0\% \\
\quad \% of trajectories with one   & 32.8\% & 33.7\% & 14.3\% & 0.0\% \\
\quad eligible versions             & 5{,}838 & 5{,}648 & 658 & 344 \\
\quad revisits observed             & 531 & 406 & 1 & 0 \\
\quad \% that beat the version returned to & 78.5\% & 63.9\% & \multicolumn{2}{c}{$n$ too small} \\
\midrule
\multicolumn{5}{l}{\emph{Setback recovery (score only)}} \\
\quad \% of trajectories with a setback & 96.8\% & 95.6\% & 83.3\% & 98.1\% \\
\quad setbacks recovered            & 79.4\% & 72.1\% & 89.0\% & 41.1\% \\
\quad recoveries that set a new best & 95.9\% & 95.2\% & 56.2\% & 100.0\% \\
\bottomrule
\end{tabular}
\end{table}

\section{Feature Reference}
\label{app:features}

Table~\ref{tab:feature_reference} lists the 19 features used by the predictive measures of \S\ref{sec:nips_empirical}, grouped as in the main text. Each feature is computed per trajectory and summarizes one aspect of how that trajectory allocates effort over its version sequence.

\begin{table}[h]
\centering
\small
\setlength{\tabcolsep}{4pt}
\begin{tabular}{p{0.24\textwidth} p{0.68\textwidth}}
\toprule
\textbf{Feature} & \textbf{One-line description} \\
\midrule
\multicolumn{2}{l}{\emph{Ensemble timing (3)}} \\
\midrule
\texttt{has\_ens}            & Indicator for whether any ensemble-blending action appears in the trajectory. \\
\texttt{ens\_late\_minus\_early} & Q5 minus Q1 share of ensemble-blending actions; positive means ensembling concentrates late. \\
\texttt{pos\_first\_ens}     & Normalised position $[0,1]$ of the first ensemble action; $1$ if no ensemble action ever fires. \\
\midrule
\multicolumn{2}{l}{\emph{Magnitude and quintile-shift (4)}} \\
\midrule
\texttt{micro\_q1}           & Q1 share of transitions labeled magnitude $=$ micro (small early edits typical of expert iteration). \\
\texttt{major\_q1}           & Q1 share of transitions labeled magnitude $=$ major (large early sweeps, rare in expert humans). \\
\texttt{opt\_q5}             & Q5 share of transitions with intent $=$ optimisation (late-stage tuning concentration). \\
\texttt{mod\_late\_minus\_early} & Q5 minus Q1 share of transitions touching the \emph{model} coarse category. \\
\midrule
\multicolumn{2}{l}{\emph{Working-mode attention shares (4)}} \\
\midrule
\texttt{mode.train}          & Share of a version's fine state-tag mass in the \emph{in-training} mode. The four working modes each union one or two coarse state phases: \emph{in-training} (training-configuration $+$ model-definition), \emph{ensemble-blending} (ensemble $+$ inference/submission), \emph{validation-and-debugging} (validation), \emph{data-and-features} (data-io $+$ feature-engineering); a mode's share is the fraction of a version's fine-tag mass falling in it. \\
\texttt{mode.ens}            & Share in the \emph{ensemble-blending} mode (ensemble $+$ inference/submission phases). \\
\texttt{mode.val}            & Share in the \emph{validation-and-debugging} mode (validation phase). \\
\texttt{mode.data}           & Share in the \emph{data-and-features} mode (data-io $+$ feature-engineering phases). \\
\midrule
\multicolumn{2}{l}{\emph{Fine actions (3, Codex mono-loop markers from \S\ref{sec:nips_emp_pca})}} \\
\midrule
\texttt{change\_weights}     & Share of transitions that re-weight ensemble members; Codex marker. \\
\texttt{add\_member}         & Share of transitions that add a model to the ensemble; Codex marker. \\
\texttt{dependency\_mgmt}    & Share of transitions on environment or dependency debugging; Codex marker. \\
\midrule
\multicolumn{2}{l}{\emph{Fine states (5, expert K-fold practice markers from \S\ref{sec:nips_emp_index})}} \\
\midrule
\texttt{fold\_averaging}     & Share of states whose code contains K-fold prediction averaging. \\
\texttt{holdout\_split}      & Share of states whose code contains an explicit hold-out split. \emph{Negative} marker: a plain hold-out in place of $K$-fold, and the only one of the six index components whose association with rank runs the opposite way (see Table~\ref{tab:index}). \\
\texttt{oof\_prediction}     & Share of states whose code generates out-of-fold predictions. \\
\texttt{blending}            & Share of states whose code contains a blending or weighted-average step over base predictions. \\
\texttt{multi\_model\_stack} & Share of states whose code contains a stacked multi-model architecture. \\
\bottomrule
\end{tabular}
\caption{The 19 trajectory features used in the predictive measures of \S\ref{sec:nips_empirical} and the per-cohort comparison of \S\ref{sec:nips_harness_experiment}. Q1 / Q5 denote the first and fifth normalized-position quintiles; "share" denotes the fraction of transitions or states (whichever applies) within the trajectory.}
\label{tab:feature_reference}
\end{table}

\subsection{The validation-and-ensembling discipline index}
\label{app:index}

Six of the 19 features carry almost all of the association between process and
final rank. Table~\ref{tab:index} ranks every feature by its per-competition
Spearman correlation with the kernel's \emph{final rank percentage}, computed on
the paired split and on the disjoint humans-only split. This is the fraction of
the leaderboard the kernel finished ahead of, so it is
\emph{smaller-is-better} (gold-medal kernels average $4.5\%$, no-medal $6.8\%$)
and a \emph{negative} $\rho$ marks a behavior associated with a \emph{better}
finish. Note the opposite orientation to the within-competition version
percentile of \S\ref{sec:nips_emp_cost}, where larger is better.

\begin{table}[h]
\centering\small
\begin{tabular}{rlrlr}
\toprule
& \multicolumn{2}{c}{paired ($n=358$, 7 comps)} & \multicolumn{2}{c}{humans-only ($n=3545$, 127 comps)} \\
\cmidrule(lr){2-3}\cmidrule(lr){4-5}
\# & feature & $\rho$ & feature & $\rho$ \\
\midrule
1 & \texttt{fold\_averaging}  & $-0.59$ & \texttt{pos\_first\_ens}   & $+0.35$ \\
2 & \texttt{holdout\_split}   & $+0.50$ & \texttt{fold\_averaging}   & $-0.35$ \\
3 & \texttt{mode.ens}         & $-0.49$ & \texttt{holdout\_split}    & $+0.34$ \\
4 & \texttt{pos\_first\_ens}  & $+0.47$ & \texttt{blending}          & $-0.33$ \\
5 & \texttt{oof\_prediction}  & $-0.43$ & \texttt{change\_weights}   & $-0.32$ \\
6 & \texttt{change\_weights}  & $-0.42$ & \texttt{mode.ens}          & $-0.31$ \\
7 & \texttt{blending}         & $-0.40$ & \texttt{add\_member}       & $-0.27$ \\
\midrule
\multicolumn{5}{l}{\emph{remaining 12 features: $|\rho| \le 0.35$ (paired) and $\le 0.25$ (humans-only)}} \\
\bottomrule
\end{tabular}
\caption{Feature-rank association: $n$-weighted mean of per-competition Spearman
$\rho$ with leaderboard percentile. Competition-cluster bootstrap CIs for the
top eight exclude zero on both splits. Five features enter the top six on both
splits; \texttt{oof\_prediction} and \texttt{blending} are near-collinear
markers of the same practice, and we carry \texttt{blending} as the
representative. \texttt{change\_weights} is kept despite also being a Codex
mono-loop marker (CIs $[-0.48,-0.35]$ paired, $[-0.37,-0.28]$ humans-only); it
is non-monotone across the full range, which is what makes it the saturated
component of \S\ref{sec:nips_harness_results}. All associations are
correlational and co-vary with trajectory length.}
\label{tab:index}
\end{table}

The six compose one construct, \emph{validation-and-ensembling discipline}:
use $K$-fold rather than a plain hold-out (\texttt{fold\_averaging} up,
\texttt{holdout\_split} down), persist out-of-fold predictions, start
ensembling early (\texttt{pos\_first\_ens} low), spend attention in the
ensemble-and-inference mode, and blend real members (\texttt{blending} up).
The harness prompt of \S\ref{sec:nips_harness} operationalizes this construct
clause by clause. It describes where human practice sits, not a quantity to
push arbitrarily far; \texttt{change\_weights} shows this most clearly, and
\S\ref{sec:nips_emp_index} reads it accordingly.

The scaffold-signature discussion of \S\ref{sec:nips_emp_pca} additionally names several fine action tags drawn from the 85-tag action vocabulary. Table~\ref{tab:action_reference} defines them. Each is a share of transitions labeled with that action within the trajectory.

\begin{table}[h]
\centering
\small
\setlength{\tabcolsep}{4pt}
\begin{tabular}{p{0.24\textwidth} p{0.68\textwidth}}
\toprule
\textbf{Action tag} & \textbf{One-line description} \\
\midrule
\multicolumn{2}{l}{\emph{Codex bookkeeping-loop markers}} \\
\midrule
\texttt{change\_weights}   & Re-weight the contributions of existing ensemble members. \\
\texttt{postprocess\_change} & Adjust a post-hoc step on predictions (rounding, clipping, calibration). \\
\texttt{data\_loading}     & Modify how data is read or assembled without changing features or model. \\
\texttt{add\_member}       & Add a model to the ensemble. \\
\midrule
\multicolumn{2}{l}{\emph{MLEvolve mutate-in-place markers}} \\
\midrule
\texttt{layer\_modification} & Change the architecture of the current model (add, remove, or resize layers). \\
\texttt{epoch\_change}     & Change the number of training epochs or the training schedule length. \\
\texttt{seed\_averaging}   & Average predictions across repeated runs with different random seeds. \\
\midrule
\multicolumn{2}{l}{\emph{Cross-family pivots (underused by both scaffolds)}} \\
\midrule
\texttt{checkpoint\_swap}   & Restart from a different saved checkpoint of a previously trained model. \\
\texttt{pretrained\_swap}  & Replace the model family or backbone with a different pretrained one. \\
\texttt{feature\_selection} & Add, drop, or reselect input features, changing the feature set the model sees. \\
\bottomrule
\end{tabular}
\caption{Fine action tags named in the scaffold-signature analysis of \S\ref{sec:nips_emp_pca}, grouped by the loop each characterizes. All are drawn from the 85-tag action vocabulary of the transition schema.}
\label{tab:action_reference}
\end{table}

\section{Harness Experiment Details}
\label{app:harness}

This appendix supports \S\ref{sec:nips_harness}: the scores behind the
percentile view (\ref{app:harness_scores}), the prompts that constitute the
intervention (\ref{app:prompts}), and what the prompt does not reach
(\ref{app:extra_figs}).

\subsection{Matched-Budget Scores and Ablations}
\label{app:harness_scores}

Figure~\ref{fig:progress} reads the harness result as a leaderboard percentile;
Table~\ref{tab:harness_scores} gives the underlying metric values, so the size of
each move is visible in the competition's own units. Every entry is the best
valid score its run reached, and all five arms run at the same twelve-hour budget
with the same backend, task prompt and grader.

Repeated same-condition runs bound the noise: the two \texttt{commonlit} harness
runs differ by $0.012$ RMSE and the two \texttt{equity} harness runs by $0.009$
C-index. Against the matched baseline and reading no difference below that scale,
five competitions improve (\texttt{gquest}, \texttt{aes2}, \texttt{hms},
\texttt{ranzcr}, \texttt{amex}), two are within noise (\texttt{commonlit},
\texttt{equity}), and none regress.

The two ablations separate the skill's content from the schedule that delivers
it. \emph{Abl-B} keeps the $30$-minute re-injection cadence but strips the
planning content, and it lands at or below the baseline everywhere; on
\texttt{hms} it is far worse ($1.377$ against a $1.050$ baseline). The cadence
alone contributes nothing. \emph{Abl-A} delivers a single content block instead
of the full skill, and it recovers much of the gain on \texttt{aes2} and
\texttt{hms} but none on \texttt{gquest} or \texttt{ranzcr}. The effect
therefore comes from the skill's content, and from more than one block of it.

\begin{table}[h]
\centering\small
\setlength{\tabcolsep}{4.5pt}
\caption{Best score reached per run at a matched twelve-hour budget. Arrows give
the metric direction. Two entries in a cell are independent runs under the same
condition. \emph{Abl-A} is a single content block; \emph{Abl-B} is the injection
cadence with no planning content.}
\label{tab:harness_scores}
\begin{tabular}{llrrrr}
\toprule
Competition & Metric & Baseline & Harness & Abl-A & Abl-B \\
\midrule
\texttt{commonlit} & RMSE $\downarrow$   & 0.510 & 0.505 / 0.517 & 0.520 & 0.512 \\
\texttt{equity}    & C-index $\uparrow$  & 0.675 & 0.670 / 0.680 & 0.672 & 0.670 \\
\texttt{gquest}    & Spearman $\uparrow$ & 0.371 & 0.429         & 0.372 & 0.388 \\
\texttt{aes2}      & QWK $\uparrow$      & 0.771 & 0.817 / 0.808 & 0.806 & 0.796 \\
\texttt{hms}       & KL $\downarrow$     & 1.050 & 0.718         & 0.795 & 1.377 \\
\texttt{ranzcr}    & AUC $\uparrow$      & 0.545 & 0.877         & 0.542 & 0.583 \\
\texttt{amex}      & Amex $\uparrow$     & 0.023 & 0.781         & ---   & 0.022 \\
\bottomrule
\end{tabular}
\end{table}

\subsection{Agent-Run Prompts}
\label{app:prompts}

Both prompt conditions in the harness experiment of \S\ref{sec:nips_harness_experiment} share the task prompt of \S\ref{app:prompts:task} below. They differ only in whether the planning skill block of \S\ref{app:prompts:skill} is expanded into the \texttt{\{\{SKILL\_BLOCK\}\}} placeholder, and whether the reminder of \S\ref{app:prompts:reminder} is re-injected every 30 minutes. Per-run variables (\texttt{\{\{HOURS\}\}}, \texttt{\{\{COMPETITION\_SLUG\}\}}, \texttt{\{\{DATA\_DIR\}\}}, \texttt{\{\{RUN\_DIR\}\}}, \texttt{\{\{CUDA\_VISIBLE\_DEVICES\}\}}, \texttt{\{\{MLEBENCH\_CACHE\}\}}) are filled by the runner before dispatch.

\subsubsection{Task Prompt (baseline and harness)}
\label{app:prompts:task}

{\small\begin{verbatim}
# Kaggle Competition Run

You are running unattended for **{{HOURS}} hours** on **{{COMPETITION_SLUG}}**.
Nobody will answer questions. Decide and proceed.

## Environment
- **Data** (already prepared): `{{DATA_DIR}}` -- inspect it yourself
  (`ls {{DATA_DIR}}`). The layout varies by competition but always includes
  `sample_submission.csv` which defines the required output format.
- **Working directory**: `{{RUN_DIR}}` -- write anything you need here
  (code, models, notes, intermediate files). Your final submission goes at
  `{{RUN_DIR}}/submission.csv` and must match `sample_submission.csv`
  exactly in columns and row count.
- **GPU**: `CUDA_VISIBLE_DEVICES={{CUDA_VISIBLE_DEVICES}}`

## Grading
Local, unlimited, not rate-limited. Use it whenever it's useful:
    mlebench grade-sample {{RUN_DIR}}/submission.csv {{COMPETITION_SLUG}}
                          --data-dir {{MLEBENCH_CACHE}}

## Goal
Push the score on `{{COMPETITION_SLUG}}` as high as you can within the time
budget. There is always something worth trying next; don't exit voluntarily
before the timer ends.

**The task ends ONLY when the {{HOURS}}-hour wall timer expires.** If your
agent framework provides a `finish`, `stop`, `terminate`, `end_task`, or
similar tool -- do not call it. If you think you're done, use the remaining
time to try another backbone / more folds / more features / TTA / ensembling.

{{SKILL_BLOCK}}

## Rules
- Do not read other humans' solutions on this machine.
- Do not read any files under
  `{{MLEBENCH_CACHE}}/{{COMPETITION_SLUG}}/prepared/private/`
  (the grader's ground-truth answer key).
- Do not search online for competition-specific solutions, winning notebooks,
  or leaderboard approaches for `{{COMPETITION_SLUG}}`. General library docs
  (PyTorch, sklearn, lightgbm, etc.) are fine.
- Always keep a valid `submission.csv` on disk.
\end{verbatim}}

In the \emph{baseline} condition, \texttt{\{\{SKILL\_BLOCK\}\}} expands to the empty string and no reminder is injected. In the \emph{harness} condition, it expands to \S\ref{app:prompts:skill} and the reminder of \S\ref{app:prompts:reminder} is re-injected every 30 minutes.

\subsubsection{Planning Skill Block (harness only)}
\label{app:prompts:skill}

{\small\begin{verbatim}
## ML Strategy

> Time budget: You have {{HOURS}} hours. Pacing: first ~25% baseline + K-fold
> + first ensemble; next ~50% iterate + blend + OOF stack; final ~25%
> finalize. If debugging dependencies past 30 min, abandon that path.

### Part 1 -- Avoid these patterns
- Do not use `train_test_split` / a single holdout. Use K-fold from version 1.
- Do not lock into a single-action loop (e.g., 5 consecutive "tweak hyperparam"
  steps). Cycle through validation -> feature -> model -> ensemble ->
  re-validate.
- Do not open with a major rewrite as v1. Start from a small working baseline
  and iterate.
- Do not waste rounds on environment / dependency debugging across many
  iterations. Batch them into one focused 20-30 min session, then move on.
- Do not sit in postprocessing tweak loops (rounding / clipping / calibration
  repeated). Set postprocessing once and move on.

### Part 2 -- Core practices
- K-fold CV with prediction averaging across folds.
- First ensemble call within the first ~25% of the time budget -- not at
  the end.
- Generate out-of-fold (OOF) predictions and stack them via a meta-learner
  (Ridge / LGBM-on-OOF) or weighted blend.
- Blend predictions from multiple models or seeds.
- Tune `sample_weight=` / loss when the leaderboard metric is stratified,
  weighted, or imbalanced relative to the default training loss.
- Keep adding ensemble members incrementally; do not stop at one model.
- Cache OOF arrays + trained model artifacts to disk so any base model can be
  revisited without retraining.
- Use multiple model families (e.g., transformer + tabular booster + Ridge
  head), not a single architecture.

### Part 3 -- Polish
- Make many small / micro-magnitude changes early; avoid big sweeps in the
  first 20%.
- Implement a custom validation metric that mirrors the leaderboard.
- Use a custom loss / objective when the default is not aligned with the
  metric.
- Average across multiple seeds for each base model.

### Part 4 -- Task-type conditional
- Text / NLP: multiple pretrained transformer backbones (DeBERTa, RoBERTa,
  Electra, ...); blend predictions; multi-seed averaging per backbone.
- Tabular: iterate on derived features -- groupby aggregations, ratios,
  time-since-event windows, target-mean encoding for categoricals. Add
  2-3 new engineered features per iteration before tweaking hyperparameters.
- Image / CV: test-time augmentation (flipped / rotated / cropped versions
  of test inputs, averaged); heavy training-time augmentation (random crops,
  color jitter, mixup).
- Time-series / signal: efficient I/O (parquet, polars); window-based
  aggregate features.

### Part 5 -- Self-check every 30-60 minutes
1. Is my validation strategy K-fold, not single train_test_split?
2. Does my CV metric match the leaderboard metric exactly?
3. Have I considered whether sample_weight= or a custom loss is needed?
4. Have I made my first ensemble call yet? If past 25% of budget, do it now.
5. Have I saved OOF predictions per base model to disk?
6. Am I blending at least 2 models or 2 seeds?
7. Have I been stuck on the same task category for 3+ consecutive steps?
\end{verbatim}}

The block totals roughly 1\,k tokens, mirroring the structure of the human-prior practices identified in \S\ref{sec:nips_emp_index} (cycling working modes, early ensemble, OOF caching, K-fold validation, model-family pivots).

\subsubsection{Periodic Reminder (harness only)}
\label{app:prompts:reminder}

Re-injected by the runner every 30 minutes after the run reaches the first ensemble window, to keep the planning rules active across long horizons.

{\small\begin{verbatim}
## Self-check (forced reminder from harness)

You have been running for a while. Before continuing your current step, run
through these 7 questions. For each "no", fix it before adding new things --
these are the strongest predictors of finishing well.

1. Is my validation strategy K-fold, not a single train_test_split?
2. Does my CV metric match the leaderboard metric exactly (same formula,
   same stratification, same weighting)?
3. Have I considered whether sample_weight= or a custom loss is needed for
   this metric?
4. Have I made my first ensemble call yet? If past ~25% of my time budget
   and the answer is "no", do it now -- even a simple 2-model average counts.
5. Have I saved OOF predictions per base model to disk?
6. Am I blending at least 2 models or 2 seeds? How many ensemble members
   do I currently have?
7. Have I been stuck on the same task category for 3+ consecutive steps?

After auditing, continue iterating on the actual solution -- do not produce
a long write-up. Run code and improve submission.csv.
\end{verbatim}}

\subsection{Behaviors the Prompt Does Not Reach}
\label{app:extra_figs}

Read on behaviors the prompt does not name directly, the harness moves
diagnosis rhythm and timing but overshoots the human band on both, while the
dynamics of how effort is allocated stay agent-like. Assigning each version to
the working mode carrying most of its fine-tag mass (\S\ref{app:features}) and
counting how often that dominant mode changes from one version to the next, the
harnessed run switches on $2.2\%$ of steps against $13.2$--$14.1\%$ for every
human cohort, close to the $7.1\%$ of prior Codex and far from any human value.
The prompt therefore reaches the practices it names without reaching the
rhythm in which a run moves between them, which is the dissociation
\S\ref{sec:nips_harness_results} reads as the boundary of what instruction
changes.

\section{Limitations}
\label{app:limitations}

Four limits bound what the data can say. Human trajectories are reconstructed from public Kaggle notebook histories, which record saved versions, not all work: participants run private local experiments, reuse public notebooks, and sometimes publish cleaned-up versions after off-platform work. The human and agent settings are also task-aligned but not fully controlled; humans work over longer calendar horizons with different tooling, compute, and collaboration. We therefore read human behavior as a reference distribution of public practice, not as optimal planning (\S\ref{sec:alignment}).

The annotations are inferred, not observed. State and action labels are grounded in code structure and diffs, but intent is inferred from the change, so we use it only at a coarse level and read it together with actions, timing, and score changes. Intent labels are not observations of what a developer was thinking. Finally, trajectory units are not independent: notebooks share fork lineage and code, and MLEvolve branches share tree nodes, which is why every interval in the paper clusters by competition, lineage, and run (Appendix~\ref{app:reuse}).

More agent harnesses, newer competitions, non-Kaggle workflows, and richer execution logs would reduce platform-specific bias; larger gold annotation sets and matched-budget reruns would tighten the label and budget caveats. \S\ref{sec:outlook} discusses what the measurements point toward.
\clearpage
\section*{NeurIPS Paper Checklist}

\begin{enumerate}

\item {\bf Claims}
    \item[] Question: Do the main claims made in the abstract and introduction accurately reflect the paper's contributions and scope?
    \item[] Answer: \answerYes{}
    \item[] Justification: The abstract and introduction state the five contributions (dataset, schema, process-level analysis, planning harness, and toolkit). Section~\ref{sec:nips_empirical} and Section~\ref{sec:nips_harness} support them, and the harness is presented as closing part of the gap, not all of it.

\item {\bf Limitations}
    \item[] Question: Does the paper discuss the limitations of the work performed by the authors?
    \item[] Answer: \answerYes{}
    \item[] Justification: Appendix~\ref{app:limitations} covers the incomplete record of off-platform human work, the task-aligned but not fully controlled human and agent settings, inferred intent labels, and non-independent trajectory units. Section~\ref{sec:alignment} frames human behavior as a reference distribution of public practice.

\item {\bf Theory assumptions and proofs}
    \item[] Question: For each theoretical result, does the paper provide the full set of assumptions and a complete (and correct) proof?
    \item[] Answer: \answerNA{}
    \item[] Justification: The paper is empirical and presents no theoretical results.

\item {\bf Experimental result reproducibility}
    \item[] Question: Does the paper fully disclose all the information needed to reproduce the main experimental results of the paper to the extent that it affects the main claims and/or conclusions of the paper (regardless of whether the code and data are provided or not)?
    \item[] Answer: \answerYes{}
    \item[] Justification: Section~\ref{sec:nips_dataset} and Appendix~\ref{app:construction} describe extraction, filtering, alignment, and annotation. Section~\ref{sec:nips_harness_experiment} and Appendix~\ref{app:harness} describe the harness protocol, including the full agent prompts (Appendix~\ref{app:prompts}).

\item {\bf Open access to data and code}
    \item[] Question: Does the paper provide open access to the data and code, with sufficient instructions to faithfully reproduce the main experimental results, as described in supplemental material?
    \item[] Answer: \answerYes{}
    \item[] Justification: The dataset, labelers, code, and project page are public at the URLs on the first page. Kaggle competition data is not redistributed, and human notebook source is redistributed only where the kernel license permits it (Section~\ref{sec:dataset_quality}).

\item {\bf Experimental setting/details}
    \item[] Question: Does the paper specify all the training and test details (e.g., data splits, hyperparameters, how they were chosen, type of optimizer) necessary to understand the results?
    \item[] Answer: \answerYes{}
    \item[] Justification: The paper specifies the competitions, cohort definitions and analysis scopes (Appendix~\ref{app:scope}), the 12-hour agent budget, backend, prompts, and grader (Section~\ref{sec:nips_harness_experiment}), and the teacher-student labeling setup with its held-out evaluation (Section~\ref{sec:dataset_quality}, Appendix~\ref{app:additional_empirical}). Labeler training hyperparameters are not listed in the paper because the findings depend on the resulting labels, whose reliability is reported directly.

\item {\bf Experiment statistical significance}
    \item[] Question: Does the paper report error bars suitably and correctly defined or other appropriate information about the statistical significance of the experiments?
    \item[] Answer: \answerYes{}
    \item[] Justification: Appendix~\ref{app:robustness} reports 95\% two-stage cluster-bootstrap intervals and a generalized estimating equation for the headline human-agent gaps. The harness study has at most two runs per condition, so Appendix~\ref{app:harness_scores} reports a noise bound from repeated runs and not a significance test.

\item {\bf Experiments compute resources}
    \item[] Question: For each experiment, does the paper provide sufficient information on the computer resources (type of compute workers, memory, time of execution) needed to reproduce the experiments?
    \item[] Answer: \answerNo{}
    \item[] Justification: The paper reports the 12-hour wall-clock budget of each agent run, but not the hardware type, memory, or total compute.

\item {\bf Code of ethics}
    \item[] Question: Does the research conducted in the paper conform, in every respect, with the NeurIPS Code of Ethics \url{https://neurips.cc/public/EthicsGuidelines}?
    \item[] Answer: \answerYes{}
    \item[] Justification: The work analyzes public Kaggle artifacts and controlled agent runs. Redistribution follows per-kernel licenses, and notebook execution outputs are stripped at extraction (Section~\ref{sec:dataset_quality}).

\item {\bf Broader impacts}
    \item[] Question: Does the paper discuss both potential positive societal impacts and negative societal impacts of the work performed?
    \item[] Answer: \answerNo{}
    \item[] Justification: The paper discusses benefits for diagnosing and improving long-horizon ML agents (Section~\ref{sec:outlook}) and handles the privacy of public notebook data (Section~\ref{sec:dataset_quality}), but it does not include a broader discussion of negative societal impacts, such as those of more capable autonomous ML agents.

\item {\bf Safeguards}
    \item[] Question: Does the paper describe safeguards that have been put in place for responsible release of data or models that have a high risk for misuse (e.g., pre-trained language models, image generators, or scraped datasets)?
    \item[] Answer: \answerYes{}
    \item[] Justification: The corpus is derived from public platform data. Section~\ref{sec:dataset_quality} describes the safeguards: per-kernel license checks before redistribution, stripping of notebook execution outputs, and no redistribution of competition data. The released labelers are narrow classifiers for this schema.

\item {\bf Licenses for existing assets}
    \item[] Question: Are the creators or original owners of assets (e.g., code, data, models), used in the paper, properly credited and are the license and terms of use explicitly mentioned and properly respected?
    \item[] Answer: \answerYes{}
    \item[] Justification: The paper cites the benchmarks, agents, and base models it uses. Section~\ref{sec:dataset_quality} states the license handling: human notebooks are checked per kernel against the Meta Kaggle Code mirror, the labeler weights inherit Apache 2.0 from Qwen3, and Kaggle competition data is not redistributed.

\item {\bf New assets}
    \item[] Question: Are new assets introduced in the paper well documented and is the documentation provided alongside the assets?
    \item[] Answer: \answerYes{}
    \item[] Justification: TraceML is released with a datasheet and Croissant metadata (Section~\ref{sec:dataset_quality}), alongside the labelers and the extraction toolkit (Appendix~\ref{app:toolkit}).

\item {\bf Crowdsourcing and research with human subjects}
    \item[] Question: For crowdsourcing experiments and research with human subjects, does the paper include the full text of instructions given to participants and screenshots, if applicable, as well as details about compensation (if any)?
    \item[] Answer: \answerNA{}
    \item[] Justification: The paper involves no crowdsourcing and recruits no participants. Human trajectories are reconstructed from public Kaggle notebook histories and metadata.

\item {\bf Institutional review board (IRB) approvals or equivalent for research with human subjects}
    \item[] Question: Does the paper describe potential risks incurred by study participants, whether such risks were disclosed to the subjects, and whether Institutional Review Board (IRB) approvals (or an equivalent approval/review based on the requirements of your country or institution) were obtained?
    \item[] Answer: \answerNA{}
    \item[] Justification: The paper has no study participants. It analyzes public platform artifacts and controlled agent runs.

\item {\bf Declaration of LLM usage}
    \item[] Question: Does the paper describe the usage of LLMs if it is an important, original, or non-standard component of the core methods in this research? Note that if the LLM is used only for writing, editing, or formatting purposes and does \emph{not} impact the core methodology, scientific rigor, or originality of the research, declaration is not required.
    \item[] Answer: \answerYes{}
    \item[] Justification: LLMs are central to the methods. The agent trajectories come from LLM-based systems (Section~\ref{sec:agent_collection}), and the annotation pipeline uses a \texttt{gpt-5.4-mini} teacher with distilled Qwen3-1.7B labelers (Section~\ref{sec:dataset_quality}).

\end{enumerate}

\end{document}